%% file: main.tex
\documentclass[runningheads]{llncs}
\usepackage[T1]{fontenc}
\usepackage{graphicx}
\usepackage{amsmath, amssymb, caption}
\usepackage[numbers]{natbib}
\usepackage{tabularx, booktabs, subcaption, multirow, makecell}
\usepackage{fontawesome5}
\usepackage{tikz}
\usetikzlibrary{decorations.pathreplacing}
\usetikzlibrary{shapes.geometric, arrows, positioning, fit}
\tikzstyle{arrow} = [thick,->,>=stealth]
\usepackage[short, nocomma]{optidef}
\allowdisplaybreaks

\begin{document}
\title{
Transit Network Design with Two-Level Demand Uncertainties: A Machine Learning and Contextual Stochastic Optimization Framework
}
%
%\titlerunning{Abbreviated paper title}
% If the paper title is too long for the running head, you can set
% an abbreviated paper title here
%
\author{
Hongzhao Guan
\inst{1} \orcidID{0000-0002-5006-1165}
\and
Beste Basciftci
\inst{2} \orcidID{0000-0002-3876-2559}
\and
Pascal Van Hentenryck
\inst{1} \orcidID{0000-0001-7085-9994}
}
\authorrunning{H. Guan et al.}
% First names are abbreviated in the running head.
% If there are more than two authors, 'et al.' is used.
%
\institute{H. Milton Stewart School of Industrial and Systems Engineering, Georgia Institute of Technology, Atlanta GA 30345, USA
\email{\{hguan7, pvh\}@gatech.edu}\\
% \url{http://www.springer.com/gp/computer-science/lncs}
\and
Tippie College of Business, University of Iowa, Iowa City IA 55242, USA\\
\email{beste-basciftci@uiowa.edu}
}
\maketitle              % typeset the header of the contribution
\begin{abstract}
Transit Network Design is a well-studied problem in the field of transportation, typically addressed by solving optimization models under fixed demand assumptions. Considering the limitations of these assumptions, this paper proposes a new framework, namely the Two-Level Rider Choice Transit Network Design (2LRC-TND), that leverages machine learning and contextual stochastic optimization (CSO) through constraint programming (CP) to incorporate two layers of demand uncertainties into the network design process. The first level identifies travelers who rely on public transit (core demand), while the second level captures the conditional adoption behavior of those who do not (latent demand), based on the availability and quality of transit services. To capture these two types of uncertainties, 2LRC-TND relies on two travel mode choice models,
that use multiple machine learning models. To design a network, 2LRC-TND integrates the resulting choice models into a CSO that is solved using a CP-SAT solver. 2LRC-TND is evaluated through a case study involving over 6,600 travel arcs and more than 38,000 trips in the Atlanta metropolitan area. The computational results demonstrate the effectiveness of the  2LRC-TND in designing transit networks that account for demand uncertainties and contextual information, offering a more realistic alternative to fixed-demand models.

\end{abstract}

\input{sections/1_Introduction}
\input{sections/2_Related_Work}
\input{sections/3_Problem_and_Method}

\input{sections/4_Settings}
\input{sections/5_Results}
\input{sections/6_Conclusion}

\vspace{-3mm}
\section*{Acknowledgments}
\vspace{-3mm}
The work was partially funded based upon work supported by the National Science
Foundation under Grant No. 2112533 and Grant No. 2434302.

\bibliographystyle{splncs04}
\bibliography{reference}

\end{document}

%% file: sections/1_Introduction.tex
\section{Introduction}
\vspace{-2mm}
Transit Network Design is an important problem in transportation and urban planning, that focuses on developing effective public transportation services. Transit agencies, which operate transit networks, may need to redesign their systems when new technologies emerge or when the urban environment undergoes changes. A common approach to re-design a transit network consists in solving an optimization problem that leverages existing transit demand, typically represented by Origin-Destination (O-D) pairs. However, this method often has difficulties to account for the changes in demand, i.e., the number of riders who will actually use the system after the network re-design. For a transit agency, relying solely on transit demand of the existing system is insufficient because the new system may draw additional riders, potentially leading to overcrowding and poor service quality. On the other hand, overestimating transit demand can result in an inefficient and costly design. The transit design can then be thought of as a game between the transit agency and riders, where the transit design influences ridership, which in turn determines the quality and cost of the system. At this point, the critical issues become how to accurately capture travel  behaviors from available data and contextual information then integrate them into optimization models.

To address these challenges, this paper provides a novel framework to model the transit network design problem, namely the Two-Level Rider Choice Transit Network Design (2LRC-TND). The 2LRC-TND framework aims at finding a network design as the equilibrium point between the transit agency and its potential riders. To find this equilibrium point, 2LRC-TND introduces a contextual stochastic optimization (CSO) approach that learns contextual probability distributions to capture rider behavior effectively.

\begin{figure}
\resizebox{0.95\columnwidth}{!}{
\begin{tikzpicture}[every node/.style={font=\small}]

%--- Box dimensions
\def\h{1cm}
\def\wA{4cm}
\def\wB{11cm}
\def\wC{12cm}

%--- Draw boxes
\draw (0,0) rectangle (\wA,\h);
\draw (\wA,0) rectangle (\wA+\wB,\h);
\draw (\wA+\wB,0) rectangle (\wA+\wB+\wC,\h);

%--- Text inside boxes
\node at (\wA/2, \h/2) {
    \fontsize{16pt}{16pt}\selectfont Core Riders
};
\node at (\wA+\wB/2, \h/2) {
    \fontsize{16pt}{16pt}\selectfont Current Transit Riders with Other Choices
};
\node at (\wA+\wB+\wC/2, \h/2) {
    \fontsize{16pt}{16pt}\selectfont Current Travelers not using Transit
};

%--- Top brace with label
\draw [decorate,decoration={brace,amplitude=8pt}]
(0,\h+0.8) -- (\wA+\wB,\h+0.8)
node[midway, yshift=15pt]{
    \fontsize{16pt}{16pt}\selectfont Current Transit Demand
};

%--- Bottom braces with labels
\draw [decorate,decoration={brace,amplitude=8pt,mirror}]
(0,-0.6) -- (\wA,-0.6)
node[midway, yshift=-16pt]{
    \fontsize{17pt}{17pt}\selectfont Core Demand for 2LRC-TND
};

\draw [decorate,decoration={brace,amplitude=8pt,mirror}]
(\wA,-0.6) -- (\wA+\wB+\wC,-0.6)
node[midway, yshift=-16pt]{
    \fontsize{17pt}{17pt}\selectfont Latent (or Potential) Demand for 2LRC-TND
};

\end{tikzpicture}
}
\caption{A Diagram depicting the demand structure in 2LRC-TND.}
\label{fig:demand_diagram}
\vspace{-2mm}
\end{figure}
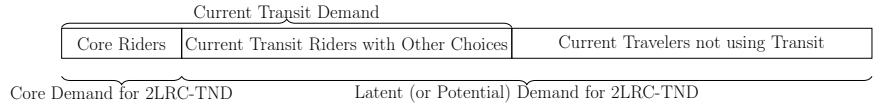
\vspace{-2mm}

More precisely, for representing travel demand, 2LRC-TND considers two levels of uncertainty in capturing core and latent demand, as shown in Figure~\ref{fig:demand_diagram}. The first level focuses on identifying  ``core demand'', i.e., travelers who rely on public transit as their primary travel mode and must be served by the transit system; these riders need to be identified from the existing transit demand. The second level captures the ``latent demand'', i.e., potential riders who may choose to use transit if the system provides good service quality; these riders have alternative travel options and will only adopt transit if the quality meets their needs. Thus, some of these potential riders are already part of the current transit demand, while others are currently rejecting the existing system and using other travel modes. By combining these two levels of uncertainty, the proposed framework introduces a novel approach to represent variable demand.

Compared to prior work in transit network design, the 2LRC-TND framework introduces novel contributions along three axes. First, it incorporates two levels of demand uncertainty; this contrasts with studies, such as \citep{guan2024path}, which divide trip demand into two fixed subsets: the core transit users and the latent demand. Importantly, 2LRC-TND treats the identification of these two sets as a classification problem. Secondly, to account for uncertainties in the network design optimization, 2LRC-TND models individual decisions as probability distributions rather than as fixed values. Lastly, 2LRC-TND is a flexible and generic framework that integrates learning and optimization in a way that is applicable to various types of network design problems beyond transit systems.

From a technical standpoint, 2LRC-TND framework leverages CSO and the Sample Average Approximation (SAA) methods to explicitly incorporate behavioral uncertainty into the transit network design optimization model. Under this setting, multiple types of machine learning techniques are used to model riders' choices. The optimization component of 2LRC-TND uses CP-SAT to maximize total system coverage while respecting budget and operational constraints.

To demonstrate the practicality of 2LRC-TND, this paper reports a comprehensive real-world case study using over 38,000 observed trips from the Atlanta metropolitan region. The results show that, within a given budget, 2LRC-TND can fully serve core trip demand while attracting additional riders, leading to a ridership increment with up to 30\% higher adoption. %and adoption rate around 30\%.
This showcases the practicality and potential social impact of 2LRC-TND for transit agencies that aim at implementing data-driven planning solutions.

In summary, the contribution of this paper can be characterized as follows:
\begin{enumerate}
    \item This paper proposes 2LRC-TND, a novel transit network design framework that captures multiple sources of uncertainty in a hierarchical way.
    \item To solve 2LRC-TND applications, the paper introduces a (widely applicable) CSO approach that integrates machine learning and stochastic optimization.
    \item Computational results from a large-scale case study with real data show that 2LRC-TND is both practical and beneficial for real-world transit network planning.
\end{enumerate}

%% file: sections/2_Related_Work.tex
\section{Related Work}
\label{sect:literature}
\vspace{-3mm}
The Transit Network Design problem has been extensively studied and is widely recognized as an essential topic for societal development \citep{guihaire2008transit, cipriani2012transit, farahani2013review}. A common approach to design a transit network is by solving an optimization problem, which is also the approach used in this paper. To help readers better understand the context of this work, it is useful to categorize existing studies based on how researchers model transit demand. Majority of studies in this field assume demand is fixed \citep{borndorfer2007column, cipriani2012transit, maheo2019benders, schobel2012line, bertsimas2021data}, while others consider uncertain or variable demand that responds to the proposed network design \citep{klier2008line, klier2015urban, canca2016general, Basciftci2020, basciftci2023capturing, guan2024path}. This paper belongs to the latter category and adopts more complex yet more realistic assumptions regarding uncertain demand by filling this critical research gap in this area.

A general approach to modeling transit demand is travel mode choice modeling, which predicts who will use transit and when, and is a key research focus in transportation. These studies primarily rely on survey data collected from individuals. Common modeling techniques range from logit models \citep{lee2018comparison} to traditional machine learning model \citep{xie2003work, zhao2020prediction}, as well as more advanced Deep Neural Networks (DNN) \citep{ma2020travel}. This paper develops its choice models by extending the insights and findings from the studies discussed above, where the existing studies in this area do not consider the subsequent network design problem while solely focusing on learning rider behavior.  

The 2LRC-TND problem explored in this paper combines the two previously discussed components---a machine learning problem and an optimization problem. Generally speaking, it is a decision-making problem that incorporates predictive models \citep{bertsimas2020predictive}, and a recently emerging framework for addressing such problems is CSO which considers contextual information and learning approaches in representing uncertainty within the optimization problems \citep{sadana2025survey}. The core ideas of these studies have been successfully applied across various domains, such as route planning in transportation \citep{tang2025enhanced}, order fulfillment \citep{ye2024contextual} and inventory control \citep{bertsimas2016inventory, wang2025data} in supply chain management, shelter location selection in disaster management \citep{jiang2025optimising}, and public school redistricting in computational social science \citep{guan2025contextual}. The next section will formally introduce the 2LRC-TND problem and establish its connection to the CSO framework.

%% file: sections/3_Problem_and_Method.tex
\section{Problem Description and Methodology}
\vspace{-2mm}
\input{sections/table_notation}
\vspace{-2mm}
This section presents the 2LRC-TND problem that integrates two levels of demand uncertainties within a stochastic optimization framework. From a technical perspective, 2LRC-TND adopts CSO as a methodological framework and employs Constraint Programming (CP) to model and solve the resulting optimization problems. The CP model aims at maximizing the expected transit coverage by making decisions on which bus arcs to activate or deactivate, while satisfying several constraints. It is important to note that the primary goal of this study is to demonstrate the practicality of the proposed CSO-based framework. The transit network design problem can be adapted to optimize other objectives, such as cost and quality of service, as long as the formulation remains compatible with the structure of the framework. Table~\ref{table:nomenclature} summarizes the notations.

\vspace{-2mm}
\subsection{Demand Uncertainty}
\vspace{-2mm}
To accurately capture demand uncertainties, two key questions must be answered. The first question identifies those riders who needs to use public transit on a daily-basis, i.e., they have no alternative travel options. Following this question, there are two demand sets that can be defined: $\mathcal{T}_{core}$ and $\mathcal{T}_{latent}$. The core set $\mathcal{T}_{core}$ captures those riders who must use transit, while the latent demand set $\mathcal{T}_{latent}$ contains those who have options. Trips in $\mathcal{T}_{latent}$ will adopt the transit system only when certain criteria are met. Together, they form the trip set $\mathcal{T}$. Therefore, the first level of demand uncertainty lies in how $\mathcal{T}$ is divided into $\mathcal{T}_{core}$ and $\mathcal{T}_{latent}$. Once the trips in $\mathcal{T}_{core}$ are identified, the transit agency must provide them with appropriate service.

The second question considers the latent set $\mathcal{T}_{latent}$ and asks when these riders choose to use transit instead of other modes?  These riders will likely make a decision on adopting or rejecting transit based on the travel path presented to them, and this is the second level of demand uncertainties. In contrast to $\mathcal{T}_{core}$, the transit agency is not obligated to serve all riders in $\mathcal{T}_{latent}$; however it aims at attracting as many of them as possible by offering them high-quality service.

With the two aforementioned questions in mind, all adoption decisions are defined as random variables coming from (unknown) distributions, which will be learned from contextual information. The context $\mathbf{x}_r$ for rider $r$ typically includes demographic, geographic, and trip-specific information. Consider two binary random variables: (1) $\mathbf{\textit{C}}_r$ which is 1 if trip $r$ is a core trip and 0 otherwise, and (2) $\mathbf{\textit{D}}_r$ is 1 if latent trip $r$ adopts the system, and 0 otherwise. The distributions of these random variables are learned by machine learning models given the contextual information. Let $U_r$ be random variable indicating whether trip $r$ utilizes the transit system, i.e.,
\begin{equation}
\label{eq:merge_two_distribution}
\resizebox{0.42\columnwidth}{!}{
$
    \mathbf{\textit{U}_r} =
    \begin{cases}
         1, \enspace if \enspace \mathbf{\textit{C}}_r = 1, \\
         1, \enspace if \enspace \mathbf{\textit{C}}_r = 0 \enspace and \enspace \mathbf{\textit{D}}_r = 1,  \\
         0, \enspace if \enspace \mathbf{\textit{C}}_r = 0 \enspace and \enspace \mathbf{\textit{D}}_r = 0,
    \end{cases}
$
}
\end{equation}

\vspace{-2mm}
\subsection{Network Design with Contextual Stochastic Optimization}
\vspace{-2mm}
This paper defines the 2LRC-TND problem over a directed multi-graph $\mathcal{G} = (\mathcal{N}, \mathcal{A})$. The set of nodes $\mathcal{N}$ includes all locations, where each $n \in \mathcal{N}$ represents a potential stop for transit service. The set of arcs $\mathcal{A}$ contains all allowed connections between pairs of locations. Each arc $a \in \mathcal{A}$ is associated with an origin, a destination, a transit mode $a_m$ such as bus and rail, and a frequency $h_a$ (for example, if the interval between two consecutive buses on this arc is 10 minutes, then $h_a$ is 6 buses per hour). The decision variable $z_a$ indicates whether arc $a \in \mathcal{A}$ is opened. The vector $\mathbf{z}$ collectively represents the network design, and the set $\mathcal{Z}$ contains all feasible network designs. Further modeling assumptions and details are discussed in subsequent sections.

2LRC-TND aims at maximizing expected transit coverage (or ridership) which can be represented in the following compact form:
\begin{equation}
\label{eq:objective_original_compact}
\max\limits_{\mathbf{z} \in \mathcal{Z}}
    \mathbb{E}_{\mathbf{U}\sim \mathbb{P}(\mathbf{U}| \mathbf{z},\mathbf{x})}
        \big[g(\mathbf{U}) \big],
\end{equation}
where the function $g(\mathbf{U})$ represents the transit coverage amount under network design $\mathbf{z}$ and  ridership $\mathbf{U}$ as defined for every trip $r$ in \eqref{eq:merge_two_distribution}, and the conditional distribution $\mathbb{P}(\mathbf{U}|\mathbf{z},\mathbf{x})$ corresponds to the adoption behavior of riders based on the all trips' context information $\mathbf{x}$ and network design $\mathbf{z}$.
The transit coverage is defined as
\begin{equation}
\resizebox{0.225\columnwidth}{!}{
$
g(\mathbf{U}) = \sum\limits_{r \in \mathcal{T}} e_r \cdot \textit{U}_r
$
}
\end{equation}

where $e_r$ is the set of riders associated with trip $r$. Assuming that each trip $r$ makes its adoption decision independently from other trips, the resulting CSO can be expressed as
\begin{equation}
\label{eq:objective_original_extensiveForm}
\resizebox{0.4\columnwidth}{!}{
$
\max\limits_{\mathbf{z} \in \mathcal{Z}}  \sum\limits_{r \in \mathcal{T}}
            e_r \cdot
    \mathbb{E}_{U_r\sim \mathbb{P}(U_r| \mathbf{z},\mathbf{x}_r)}
        [U_r].
$
}
\end{equation}
\vspace{-1mm}

2LRC-TND uses the Sample Average Approximation (SAA) method \citep{kleywegt2002sample}  to approximate the CSO problem  \eqref{eq:objective_original_extensiveForm}.
This approach generates $I$ independent and identically distributed scenarios by sampling from the distribution $\mathbb{P}(\mathbf{\textit{U}}_r | \mathbf{z}, \mathbf{x}_r)$.

Let $\text{SAA}_r^i(\mathbf{z})$ represents the adoption decision made by trip $r \in \mathcal{T}$ under the network design $\mathbf{z}$ in scenario $i \in I$. Then, the 2LRC-TND problem in \eqref{eq:objective_original_extensiveForm} can be reformulated as follows:
\begin{equation}
\resizebox{0.33\columnwidth}{!}{
$
    \max\limits_{\mathbf{z} \in \mathcal{Z}}
    \cfrac{1}{I}
    \sum\limits_{i =  1}^I
    \sum\limits_{r \in \mathcal{T}}
        e_r \cdot \text{SAA}_r^i(\mathbf{z}).
$
}
\end{equation}
\vspace{-2mm}

\subsection{Decisions and Paths}
\label{subsect:decisions_and_paths}
\vspace{-1mm}
To estimate $\mathbb{P}(\mathbf{\textit{U}}_r  | \mathbf{z}, \mathbf{x}_r)$, 2LRC-TND uses probabilistic choice models $\mathcal{C}_\mathrm{core} (\mathbf{x}_r)$ and $\mathcal{C}_\mathrm{adopt} (\mathbf{x}_r, p)$ that capture the distributions of $C_r$ and $D_r$, respectively. In general, both $\mathcal{C}_\mathrm{core}$ and $\mathcal{C}_\mathrm{adopt}$ are trained machine learning models based on the contextual information and transit paths suggested to potential riders.
Both models capture the probabilities associated with the positive labels. Then based on the probabilities, model $\mathcal{C}_\mathrm{core}$ and $\mathcal{C}_\mathrm{adopt}$ can be further used for classification task, where $\mathcal{C}_\mathrm{core}$ decide whether a trip $r$ is a core trip (label 1) or a latent trip (label 0) and $\mathcal{C}_\mathrm{adopt}$ classifies whether a latent trip $r$ would adopt (1) or reject (0) a given transit path $p$.  Following this, $C_r$ can be modeled as:
\begin{equation}
\label{eq:core_paths}
\resizebox{0.33\columnwidth}{!}{
$
    \mathbf{\textit{C}}_r =
    \begin{cases}
        1, \enspace \text{if}  \enspace  \mathcal{C}_\mathrm{core} (\mathbf{x}_r) = 1, \\
        0, \enspace \text{otherwise.}
    \end{cases}
$
}
\end{equation}

To model $\mathcal{C}_\mathrm{adopt}$, define a transit path $p$ as a route that links the origin and destination of a trip through transit services. Each path $p$ typically includes two walking trip-legs and one or more transit trip-legs. Consider the network $\mathbf{z}$ where all the arcs (e.g., bus routes) are opened, and define, for each trip $r \in \mathcal{T}$, the set $\mathcal{P}_r$ representing the $w_r$ paths with the shortest travel times, where $w_r$ is a small number. For each trip $r \in \mathcal{T}_{latent}$, the choice model $\mathcal{C}_\mathrm{adopt}(\mathbf{x}_r ,p)$ takes two inputs, the contextual information $\mathbf{x}_r$ and information associated with a transit path $p$ (e.g., the number of transfers and the transit time); it outputs the adoption decision for trip $r$. The intuition behind $\mathcal{C}_\mathrm{adopt}$ is that individuals decide whether to use transit based on the availability of a favorable path. With these definitions, $D_r$ is modeled as:

\begin{equation}
\label{eq:adopting_paths}
\resizebox{0.63\columnwidth}{!}{
$
    \mathbf{\textit{D}}_r =
    \begin{cases}
        1, \enspace \text{if} \bigvee\limits_{p \in \mathcal{P}_r}
        (feasible(\mathbf{z},p) \enspace \wedge \enspace  \mathcal{C}_\mathrm{adopt} (\mathbf{x}_r, p) = 1), \\
        0, \enspace \text{otherwise}
    \end{cases}
$
}
\end{equation}

\noindent
Here, $feasible(\mathbf{z},p)$ holds if path $p$ is feasible under the network design $\mathbf{z}$, which means if all transit trip-legs in path $p$ are open. Therefore, following Equation~\eqref{eq:adopting_paths}, trip $r$ selects transit under design $\mathbf{z}$ if there exists a feasible path $p$ that $r$ adopts. Note that, for latent trips, riders are assumed to consider only the paths in $\mathcal{P}_r$, as these represent the fastest options. Any path with longer travel times are treated as rejections.

Both $\mathcal{C}_\mathrm{core}$ and $\mathcal{C}_\mathrm{adopt}$ are treated as black-boxes within the 2LRC-TND framework and can range from complex models, such as deep neural networks, to simple rule-based approaches. Importantly, under the black-box setting, even without high-quality or sufficient data to construct detailed contextual features $\mathbf{x}_r$, one can still develop reasonable rule-based models based on domain expertise and experiences to make the 2LRC-TND framework function effectively.

The overall 2LRC-TND framework can thus be summarized as follows. Given the set of trips $\mathcal{T}$, for each scenario, the first step identifies the core riders. Each core trip $r$ must be served, which means at least one path in $\mathcal{P}_r$ is  available under the network design. The remaining trips are classified as latent in that scenario. For each latent trip,  2LRC-TND checks whether any path in $\mathcal{P}_r$ is available
and adopted. If not, trip $r$ continues using other travel modes.
\vspace{-2mm}
\subsection{The Constraint Programming Model}
\vspace{-2mm}
\input{sections/formulation_arc_design}

This study adopts CP as the modeling approach due to its strengths in handling complicated logical relationships. The proposed model includes many AND and OR operations, which would require extensive linearization if implemented using Mixed-Integer Programming (MIP), significantly increasing the complexity of the model. Building on the concepts introduced earlier, Figure~\ref{fig:cp_coverage} presents the CP model. The objective function~\eqref{eq_cp:obj} sums over the $I$ SAA scenarios to compute the expected objective, where $u_r^i$ denotes the final adoption decision of trip $r$ under scenario $i$.

The first group of constraints link $z_a$ with the adoption decisions.  Constraints~\eqref{eq_cp:riders_use_transit}~and~\eqref{eq_cp:riders_adopt_feasible_paths}
implement Equations~\eqref{eq:merge_two_distribution}~and~\eqref{eq:adopting_paths}, where $c_r^i$ and $d_{r, p}^i$ are sampled, using the choice functions $\mathcal{C}_\mathrm{core}(\mathbf{x}_r)$ and $\mathcal{C}_\mathrm{adopt}(\mathbf{x}_r, p)$, respectively.
The definition  of $\mathcal{T}^i_{latent}$ and $\mathcal{T}^i_{core}$ are based on $c^i_r$: $\mathcal{T}^i_{latent} = \{r \in \mathcal{T}: c^i_r = 0\}$ and $\mathcal{T}^i_{core} = \{r \in \mathcal{T}: c^i_r = 1\}$. Constraints~\eqref{eq_cp:feasible_path_1}~and~\eqref{eq_cp:feasible_path_2} together define $f_{p}$ dynamically based on $\mathbf{z}$, which is a linearized version of $feasible(\mathbf{z}, p)$. Constraint~\eqref{eq_cp:feasible_path_1} ensures that path $p$ is infeasible if any of its transit arcs is closed. Constraint~\eqref{eq_cp:feasible_path_2} ensures that path $p$ is set as feasible when all of its transit arcs are open. Using $f_p$, Constraint~\eqref{eq_cp:must_serve_core} ensures that at least one path is feasible for each core trip, thereby guaranteeing that the core riders have access to the service. At this stage, solving the model does not explicitly determine which path each trip will use; however, the solution reveals the set of feasible path options available to each trip and identifies which latent trips adopt the new system.

Constraints~\eqref{eq_cp:budget}--\eqref{eq_cp:one_frequency_per_arc_per_mode} focus on the modeling of the transit network itself and incorporate several realistic assumptions commonly found in transit network design. It is important to note that these constraints are independent of the SAA scenarios. First, Constraint~\eqref{eq_cp:budget} ensures that the network design remains within the operational budget $B$, where $s_a$ denotes the cost of operating arc $a$ over the planning horizon (e.g., a typical weekday). Secondly, Constraint~\eqref{eq_cp:fixed_arcs} enforces that all arcs in the set $\mathcal{A}_{fixed}$ remain open. The set $\mathcal{A}_{fixed}$ typically represents existing infrastructure that the transit agency intends to preserve in the new system, such as underground subway lines. Next, Constraint~\eqref{eq_cp:flow_balance} ensures the flow balance for each node in $\mathcal{N}$ and each mode in $\mathcal{M}$, where $\mathcal{A}_{n,m}^+$ and $\mathcal{A}_{n,m}^-$ include arcs that go in and out of node $n$ with mode $m$, respectively. In this paper, $\mathcal{M}$ by default is $\{bus, rail\}$; however, one can easily customize it. Lastly, Constraint~\eqref{eq_cp:one_frequency_per_arc_per_mode} assures that any two locations $n_1$ and $n_2$ are only connected by one arc per mode, where $\mathcal{A}_{n_1, n_2, m}$ includes all arcs that start from $n_1$ to $n_2$ with mode $m$.
Note that the output consists of a set of connected arcs between nodes, whereas in real-world settings, travelers use complete bus lines rather than navigating arc by arc. Post-processing methods can be employed to convert the selected arcs into operational bus lines.

%% file: sections/table_notation.tex
\begin{table}[!t]%
    \centering%
    \scriptsize
    \renewcommand{\arraystretch}{0.9}
    \caption{Notations}
    \label{table:nomenclature}
    \vspace{-1mm}
    \begin{tabularx}{\columnwidth}{l X}%
    \toprule
    \textbf{Notation} & \textbf{Definition} \\
    \midrule
    \textbf{Key Terms}: &  \\
    Rider & A person who takes transit services. \\
    Trip $r$ & Defined by an O-D pair and a group of $e_r$ riders who travel  together and share the same travel decisions. \\
    \textbf{Sets}: &  \\
    $\mathcal{T}$ & The set of all trips (O-D Pairs).  \\
    $\mathcal{T}_{core}^i$, $\mathcal{T}_{latent}^i$ &  The partition of set $\mathcal{T}$ into $\mathcal{T}_{core}$ and $\mathcal{T}_{latent}$ under each scenario $i$. \\ 
    $\mathcal{N}$ & The set of locations for potential transit stops. \\
    $\mathcal{M}$ & The set of transit modes. Default: $\{rail, bus\}$. \\
    $\mathcal{P}_r$ & The set of considered paths for trip $r$. \\
    $\mathcal{A}$, $\mathcal{A}_p$ & The set of all arcs and arcs belong to path $p$. \\
    $\mathcal{A}_{fixed}$ & The set of all fixed arcs. \\
    $\mathcal{A}_{n,m}^+$, $\mathcal{A}_{n,m}^-$ & The set of out-arcs and in-arcs for node $n$ with mode $m$. \\
    $\mathcal{A}_{n_1, n_2, m}$ & The set of arcs from node $n_1$ to $n_2$ with mode $m$. \\
    $\mathcal{Z}$ & The set containing feasible transit network designs. \\
    %%%%%%%%%%%%%%%%%%%%%%%%%%%%%%%%%%%%%%%%%%%%%%%%%%%%%%%%%%%%%%%%%%%%%%
    \textbf{Decision Vars.}: &  \\
    $z_a$ & Binary variable  indicating if transit arc $a$ is opened in the network design. \\
    $u_r^i$ & Binary variable indicating if trip $r$ uses the transit system, under scenario $i$. \\
    $d_r^i$ & Binary variable indicating if trip $r$ adopts the transit system, under scenario $i$. \\
    $f_p$ & Binary variable indicating if path $p$ is feasible, dynamically based on $\mathbf{z}$. \\
    %%%%%%%%%%%%%%%%%%%%%%%%%%%%%%%%%%%%%%%%%%%%%%%%%%%%%%%%%%%%%%%%%%%%%% 
    \textbf{Parameters}: & \\
    $I$, $I'$ & Number of SAA scenarios for solving and evaluation, respectively. \\
    $B$ & Budget of transit agency for a planning horizon (e.g., a day). \\
    $e_r$ & Number of riders in trip $r$ who share the same travel behaviors. \\
    $s_a$ & Cost of operating arc $a$. \\
    $h_a$ & Number of buses in an hour of arc $a$. \\
    $c_r^i$ & Indicating if trip $r$ is a core trip under scenario $i$.\\
    $d_{r, p}^i$ & Indicating the adoption decision of trip $r$ on path $p$ under scenario $i$. \\
    $l_r$ & Penalty term when evaluating a network design for not serving a trip $r$. \\
    $w_r$ & Amount of paths that are considered for trip $r$. \\
    %%%%%%%%%%%%%%%%%%%%%%%%%%%%%%%%%%%%%%%%%%%%%%%%%%%%%%%%%%%%%%%%%%%%%%
    \textbf{Vectors}: & \\
    $\mathbf{z}$ & Vector of $z_a$ over arcs $a \in \mathcal{A}$, representing a transit network design. \\
    $\mathbf{x}_r$ & Contextual information vector for trip $r$. \\
    \bottomrule
    \end{tabularx}
\vspace{-3mm}
\end{table}

%% file: sections/formulation_arc_design.tex
\begin{figure}[!t]
\begin{maxi!}
    %%%%%%%%%%%%%%%%%%%%%%%%%%%%%%%%%%%%%%%%%%%%%%%%%%%%%%%%%%%%%%%%%%%%
    {}
    { \cfrac{1}{I}
        \sum\limits_{i =  1}^I \sum\limits_{r \in \mathcal{T}}
        e_r \cdot u_r^i
        \label{eq_cp:obj}
    }
    {\label{formulation:cp_coverage}}
    {}
    %%%%%%%%%%%%%%%%%%%%%%%%%%%%%%%%%%%%%%%%%%%%%%%%%%%%%%%%%%%%%%%%%%%%
    \addConstraint
    {u_r^i}
    {   
         = 
         \begin{cases}
             1 & \enspace \forall r \in \mathcal{T}_{core}^i, i \in [I]  \\
             d_r^i & \enspace \forall r \in \mathcal{T}_{latent}^i, i \in [I] \\
         \end{cases}
        \label{eq_cp:riders_use_transit}
    }
    {}
    %%%%%%%%%%%%%%%%%%%%%%%%%%%%%%%%%%%%%%%%%%%%%%%%%%%%%%%%%%%%%%%%%%%%
    \addConstraint
    {d_r^i}
    {
        = 
        \bigvee\limits_{p \in \mathcal{P}_r} 
        (f_p \wedge d_{r, p}^i)
        \enspace
        \forall r \in \mathcal{T}_{latent}^i, i \in [I]
        \label{eq_cp:riders_adopt_feasible_paths}
    }
    {}
    %%%%%%%%%%%%%%%%%%%%%%%%%%%%%%%%%%%%%%%%%%%%%%%%%%%%%%%%%%%%%%%%%%%% 
    \addConstraint
    {z_a}
    {
        \geq  f_{p}
        \enspace \forall a \in \mathcal{A}_p,
         p \in \mathcal{P}_r, r \in \mathcal{T}
         \label{eq_cp:feasible_path_1}
    }
    {}
    %%%%%%%%%%%%%%%%%%%%%%%%%%%%%%%%%%%%%%%%%%%%%%%%%%%%%%%%%%%%%%%%%%%% 
    \addConstraint
    {f_{p}}
    {
        \geq 
        \sum\limits_{a \in \mathcal{A}_p} z_a - |\mathcal{A}_p| + 1
        \enspace \forall p \in \mathcal{P}_r, r \in \mathcal{T}
        \label{eq_cp:feasible_path_2}
    }
    {}
    %%%%%%%%%%%%%%%%%%%%%%%%%%%%%%%%%%%%%%%%%%%%%%%%%%%%%%%%%%%%%%%%%%%% 
    \addConstraint
    {c_r^i}
    {
        \leq 
        \sum\limits_{p \in \mathcal{P}_r} f_p \enspace \forall r \in \mathcal{T}_{core}^i, i \in [I]
        \label{eq_cp:must_serve_core}
    }
    {}
    %%%%%%%%%%%%%%%%%%%%%%%%%%%%%%%%%%%%%%%%%%%%%%%%%%%%%%%%%%%%%%%%%%%%
    \addConstraint
    {B}
    {
        \geq \sum\limits_{a \in \mathcal{A}} s_a \cdot z_a
        \label{eq_cp:budget}
    }
    {}
    %%%%%%%%%%%%%%%%%%%%%%%%%%%%%%%%%%%%%%%%%%%%%%%%%%%%%%%%%%%%%%%%%%%%
    \addConstraint
    {z_a}
    {
        = 1 \enspace \forall a \in \mathcal{A}_{fixed}
        \label{eq_cp:fixed_arcs}
    }
    {}
    %%%%%%%%%%%%%%%%%%%%%%%%%%%%%%%%%%%%%%%%%%%%%%%%%%%%%%%%%%%%%%%%%%%%
    \addConstraint
    {
        0
    }
    {
        =
        \sum\limits_{a \in \mathcal{A}^+_{n, m}} h_a \cdot z_a
        -
        \sum\limits_{a \in \mathcal{A}^-_{n, m}} h_a \cdot z_a
        \label{eq_cp:flow_balance}
    }
    {}
    \addConstraint
    {}
    {
        \enspace \enspace \enspace 
        \forall n \in \mathcal{N}, m \in \mathcal{M}
        \notag
    }
    {}
    %%%%%%%%%%%%%%%%%%%%%%%%%%%%%%%%%%%%%%%%%%%%%%%%%%%%%%%%%%%%%%%%%%%%
    \addConstraint
    {1}
    {   
         \geq \sum\limits_{a \in \mathcal{A}_{n_1, n_2, m}} z_a \enspace 
         \forall n_1, n_2 \in \mathcal{N}, m \in \mathcal{M}
         \label{eq_cp:one_frequency_per_arc_per_mode}
    }
    {}
    %%%%%%%%%%%%%%%%%%%%%%%%%%%%%%%%%%%%%%%%%%%%%%%%%%%%%%%%%%%%%%%%%%%%
    \addConstraint
    {
        z_a
    }
    {
         \in \{0, 1\}
         \enspace \forall a \in \mathcal{A}
    }
    {}
    \addConstraint
    {
        u_r^i
    }
    {
        \in \{0, 1\}
        \enspace \forall r \in \mathcal{T}, i \in [I]
    }
    {}
    \addConstraint
    {
        d_r^i
    }
    {
        \in \{0, 1\}
        \enspace \forall r \in \mathcal{T}_{latent}^i, i \in [I]
    }
    {}
    \addConstraint
    {
        f_{p}
    }
    {
        \in \{0, 1\}
        \enspace \forall p \in \mathcal{P}_r, r \in \mathcal{T}
    }
    {}
\end{maxi!}
\caption{The Constraint Programming Model.}
\label{fig:cp_coverage}
\vspace{-5mm}
\end{figure}

%% file: sections/4_Settings.tex
\vspace{-4mm}
\section{The Atlanta Case Study}
\label{sect:settings}
\vspace{-3mm}
This section presents a case study based in Atlanta, Georgia, USA. In summary, by solving a large-scale 2LRC-TND problem, this case study redesigns a transit system with 38,179 trips ($\mathcal{T}$) in the city of Atlanta, where services are currently operated by the Metropolitan Atlanta Rapid Transit Authority (MARTA). To ease the processes of constructing meaningful datasets, the entire case study is conducted within a rectangular area (see Figure~\ref{subfig:rail_and_rectangle}) that closely aligns with the I-285 interstate and MARTA's service area. Figure~\ref{fig:flowchart} illustrates how the case study fits within the 2LRC‑TND framework. This case study utilizes three datasets: \textsc{Survey-Arc} for training  $\mathcal{C}_\mathrm{core}$, \textsc{L-Arc-9} for training $\mathcal{C}_\mathrm{adopt}$, and \textsc{L-Arc-8} for evaluating the complete 2LRC-TND framework.

\input{sections/case_study_flowchart}

\vspace{-2mm}
\subsection{Construct $\mathcal{C}_\mathrm{core}$ and $\mathcal{C}_\mathrm{adopt}$}
\vspace{-1mm}
The model $\mathcal{C}_\mathrm{core}$ predicts the probability of a trip $r$ being a core trip. It is trained on the \textsc{Survey-Arc} dataset, which is constructed from a 2019 transit rider survey conducted by the Atlanta Regional Commission (ARC) \citep{arc2020survey}. The ground-truth labels contain 13,607 core trips (class 1) and 5,233 transit trips with other choices (class 0), where each trip represents a single rider. Since no prior study has directly addressed core trip classification, this paper adapts this survey by defining core trips as those made by commuters (trips between home and works) who cannot travel without transit, use transit at least twice weekly, and walk to/from their stops. All other commuters in the dataset are classified to class 0. The contextual features include demographic characteristics (e.g., age, income level), geographic data (e.g., number of bus stops within 5 minutes walking distance from the trip origin), and travel time information (e.g., travel time from origin to destination if driving a car).

A separate dataset, referred as \textsc{L-Arc-9}, is used to train the $\mathcal{C}_{\mathrm{adopt}}$ model. The \textsc{L-Arc-9} dataset is derived from ARC's 2020 activity-based simulation for the Atlanta metropolitan area \citep{arc2025activity}, detailed data processing procedures are documented in the Appendix of \cite{guan2022heuristic}. The original simulation study covers 5 million people and 13 million tours in a regular workday, from which \textsc{L-Arc-9} selects 15,109 home-to-work trips made between 9-10 AM that overlaps the rectangular area in Figure~\ref{subfig:rail_and_rectangle}. This highly imbalanced dataset contains 1,019 adopting trips (class 1) and 14,090 rejecting trips (class 0). Adopting trips are made by commuters who use transit despite having spare vehicles at home, while rejecting trips are made by those who drive alone despite transit availability in their area. Each trip represents a single rider and includes demographic, geographic, and trip-related contextual features. Since no existing datasets or surveys directly capture adoption behavior, this dateset provides a practical alternative to resource-intensive data collection.

In summary, the \textsc{Survey-Arc} and \textsc{L-Arc-9} datasets are used solely for training the choice models. Although the datasets originate from different sources, both were collected by the same agency within the same geographic region. Additionally, to ensure the trips under study follow relatively fixed patterns, the analysis focuses on commuting trips, as they best represent consistent travel behaviors on a regular day.
\vspace{-1mm}
\subsection{Demand Set $\mathcal{T}$, Network Arcs, and Paths}
\vspace{-1mm}
A third dataset, \textsc{L-Arc-8}, is introduced here to represent the demand used for evaluating the 2LRC-TND framework. A separate evaluation dataset is
necessary to assess the framework's performance on unseen data and avoid overfitting from the training sets used for the choice models. This dataset is derived from the same ARC simulation source as \textsc{L-Arc-9} but focuses on all commuters traveling in the rectangular area during 8-9 AM, the peak morning rush hour.

There are 38,179 trips ($\mathcal{T}$) in \textsc{L-Arc-8}, with each trip having a single rider ($e_r = 1$). Among these trips, 31,520 commute by driving alone, while 6,659 are current transit trips. All trips have contextual information compatible with the trained $\mathcal{C}_{\mathrm{core}}$ and $\mathcal{C}_{\mathrm{adopt}}$ models. Following the steps illustrated in Figure~\ref{fig:flowchart}, for each SAA scenario, the 6,659 existing transit trips are first split into core trips
and current adopting trips using $\mathcal{C}_{\mathrm{core}}$. The current adopting trips and the 31,520 driving trips then form the latent demand for 2LRC-TND.

This case study assumes removal of all existing bus lines while preserving the rail lines (see Figure~\ref{subfig:rail_and_rectangle}), as rail infrastructure is not flexible and difficult to modify. Within the defined rectangular area, there are 933 stops ($\mathcal{N}$), including 38 rail stations. To construct the set $\mathcal{A}$ and keep the CP model computationally manageable, when considering potential bus connection for each stop, only the five nearest stops based on driving distance are considered. For each potential connection between two stops, a fixed frequency of six buses per hour is assumed ($h_a = 6$ for all arcs). In addition to the standard on-road travel time, an extra five minutes is added to account for peak-hour congestion, waiting time, and other potential delays. In total, the network includes 6,004 arcs ($\mathcal{A}$), of which 692 are fixed rail arcs ($\mathcal{A}_{fixed}$). To model the operating cost between 8 a.m. and 9 a.m., the case study assumes a rate of \$72.15 per hour for buses while in motion. Rail operating costs are dropped from the budget $B$, as all rail arcs are fixed and not subject to redesign. The cost estimation and stops information can be found in an Atlanta case study \citep{auad2021resiliency}.

To enumerate the paths, a graph is first constructed considering all arcs in $\mathcal{A}$ are open. For each trip $r$, the origin and destination are connected to their five nearest stops based on walking distance, and the four $(w_r)$ fastest paths are then selected using shortest-path algorithms to construct the set $\mathcal{P}_r$. $w_r$ is set to 4 to maintain reasonable model size. Additionally, real-world navigation apps commonly considers a similar number of transit paths for users. When constructing SAA scenarios $i$, an adoption decision ($d_{r,p}^i$) for each path in $\mathcal{P}_r$ is sampled based on the predicted adoption probability from the pre-determined model $\mathcal{C}_{adopt}$. Similarly, $c_r^i$ are sampled using the predicted probability from $\mathcal{C}_\mathrm{core}$. After sampling, for latent trips where $c_r^i = 0$, if all paths in $\mathcal{P}_r$ are rejecting paths, these trips can be excluded from the optimization model since these trips will reject the network regardless.
\vspace{-1mm}
\subsection{Experiments: Choice Modeling and Evaluating 2LRC-TND}
\vspace{-1mm}
The first group of experiments in this case study focuses on evaluating the  performance of $\mathcal{C}_\mathrm{core}$ and $\mathcal{C}_\mathrm{adopt}$ when treated as classifiers. Specifically, $\mathcal{C}_\mathrm{core}$ is evaluated on the \textsc{Survey-Arc} dataset, while $\mathcal{C}_\mathrm{adopt}$ is evaluated on \textsc{L-Arc-9}. Three types of machine learning models are used: Multinomial Logit (L), Random Forest (RF), and Deep Neural Network (DNN). The dataset is randomly split using an 80-20 training-test split. This process repeats 100 times.

For each training-test split, hyper-parameters are tuned via a grid-search 5-fold cross-validation of the training set, and the best configuration is evaluated on the test set. Specifically, for the Logit model, the regularization parameter $C$ is tested with values \{0.1, 0.8, 1.0, 10\}, maximum iterations with \{100, 500, 2000\}, and class weight ratios of \{1:2, 1:3, 1:5, 1:10\} are evaluated for $\mathcal{C}_{\text{adopt}}$ to address class imbalance. The Random Forest models test the number of estimators with values \{500, 1000, 2000\} and maximum tree depth with \{10, 15, 20, 25\}, with the class weight ratios of \{1:2, 1:3, 1:5, 1:10\}  also explored for $\mathcal{C}_{\text{adopt}}$.  The DNN architecture employs embeddings for categorical and ordinal variables, followed by two fully connected layers with 32 and 16 neurons respectively, each using a dropout rate of 0.2 and the Adam optimizer. Learning rates are tuned with candidate values \{0.001, 0.01, 0.1, 0.15\}, and early stopping is applied with a patience of 20 epochs and tolerance of 0.001.

The second group of experiments evaluates the 2LRC-TND framework by comparing it against three benchmarks. The first benchmark, Fixed-Demand (FD), designs a network using only the 6,659 existing transit riders in a deterministic manner, without considering demand uncertainties. The second benchmark, Naive Rule-Based (RB), with no involvement of Machine Learning, uses hand-crafted rules to perform the same functions as $\mathcal{C}_{\mathrm{core}}$ and $\mathcal{C}_{\mathrm{adopt}}$. A simple rule-based model example is: "If both the origin and destination of a trip are within a 5-minute walk distance to a transit station, assign a core trip probability of 0.8; otherwise, assign 0.2." In general, a rule-based approach requires extensive domain expertise, but the design of such rules is beyond the scope of this paper. 
The third benchmark, Deterministic, employs the same machine learning models as 2LRC-TND but considers them under a fixed threshold of 0.5 to obtain deterministic decisions and designs the network under this setting.

For each of these runs (2LRC-TND and Benchmarks), two budget levels are tested. These values are chosen based on the total cost of activating all available arcs from 8AM to 9AM, which is approximately \$58K. A budget of \$30K represents roughly half of this total, while \$35K corresponds to 60\%, allowing for a meaningful comparison under different budget constraints.

All implementations for this paper are programmed in Python 3.11, with the support of machine learning packages PyTorch, Scikit-learn, and imblearn. The CP-SAT solver from OR-Tools is used for solving the optimization model, configured to run with 8 threads with 12 hours solving time. All experiments were conducted using the CPUs and RTX-6000 GPUs available on a Linux High Performance Computing cluster.
\vspace{-1mm}
\subsection{Evaluation Metrics for Network Designs}
\vspace{-1mm}
Network designs can be evaluated by standard transportation metrics, such as the number of people covered and the best travel time offered to riders. This study presents an additional evaluation approach based on SAA,
where a network design is assessed using out-of-sample scenarios (denoted as $I'$ and $I' >> I$), different than $I$ scenarios used during network design. Since a network is optimized only with respect to a specific set of scenarios, it may not be feasible for scenarios outside that set, i.e., Constraint~\eqref{eq_cp:must_serve_core} might not be satisfied. Function $Eval$ evaluates a network design $\mathbf{z}$ by quantifying the number of riders it successfully serves and by penalizing the core trips $r$ with no feasible paths with cost $l_r$.  The resulting $Eval$ function can be represented as follows by considering the optimization model in Figure~\ref{fig:cp_coverage} under the given network design $\mathbf{z}$ over $I'$ out-of-sample scenarios:
\begin{equation}
\resizebox{0.60\columnwidth}{!}{
$
Eval(\mathbf{z}) =
    \cfrac{1}{I'}
    \sum\limits_{i =  1}^{I'} \sum\limits_{r \in \mathcal{T}}
    e_r
    \cdot
    (
        u_r^i -
        l_r
        \cdot
        \max(c^i_r-\sum\limits_{p \in \mathcal{P}_r} f_p,0)
    )
$
}
\end{equation}

%% file: sections/case_study_flowchart.tex
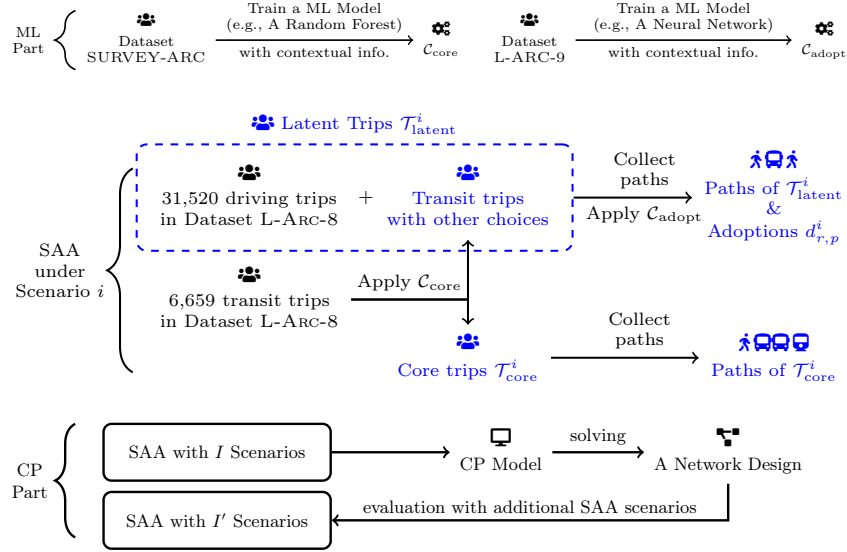
\begin{figure}
% --- --- --- --- --- --- --- --- --- --- Construct C_core
\begin{subfigure}[b]{0.99\columnwidth}
\resizebox{0.95\columnwidth}{!}{
\begin{tikzpicture}[thick, font=\scriptsize]

\draw [decorate,decoration={brace,amplitude=10pt,mirror}]
        (-1.2,0.5) -- (-1.2,-0.5) node [midway,xshift=-0.8cm, align=center] {ML \\ Part};

\node (data_survey) at (0,0) {%
  \begin{tabular}{c}
    \faUsers \\[2pt] Dataset \\ \textsc{SURVEY-ARC}
  \end{tabular}
};

\node (model_core) at (5,0) {%
  \begin{tabular}{c}
    \faCogs \\[2pt] \textsc{$\mathcal{C}_{\mathrm{core}}$}
  \end{tabular}
};

\draw[->] (data_survey) -- node[above, align=center]{Train a ML Model \\ (e.g., A Random Forest)} node[below, align=center]{with contextual info.} (model_core);

\node (data_survey) at (6.5,0) {%
  \begin{tabular}{c}
    \faUsers \\[2pt] Dataset \\ \textsc{L-ARC-9}
  \end{tabular}
};

\node (model_adopt) at (11.5,0) {%
  \begin{tabular}{c}
    \faCogs \\[2pt] \textsc{$\mathcal{C}_{\mathrm{adopt}}$}
  \end{tabular}
};

\draw[->] (data_survey) -- node[above, align=center]{Train a ML Model \\ (e.g., A Neural Network)} node[below, align=center]{with contextual info.} (model_adopt);
\end{tikzpicture}
}

\end{subfigure}
% --- --- --- --- --- --- --- --- --- --- --- --- --- --- --- --- --- --- --- ---
% --- --- --- --- --- --- --- --- --- --- --- --- --- --- --- --- --- --- --- ---
% --- --- --- --- --- --- --- --- --- --- --- --- --- --- --- --- --- --- --- ---
% --- --- --- --- --- --- --- --- --- --- --- --- --- --- --- --- --- --- --- ---
% --- --- --- --- --- --- --- --- --- --- --- --- --- --- --- --- --- --- --- ---
\vskip\baselineskip
% --- --- --- --- --- --- --- --- --- --- --- --- --- --- --- --- --- --- --- ---
% --- --- --- --- --- --- --- --- --- --- --- --- --- --- --- --- --- --- --- ---
% --- --- --- --- --- --- --- --- --- --- --- --- --- --- --- --- --- --- --- ---
% --- --- --- --- --- --- --- --- --- --- --- --- --- --- --- --- --- --- --- ---
% --- --- --- --- --- --- --- --- --- --- --- --- --- --- --- --- --- --- --- ---
\begin{subfigure}[b]{0.99\columnwidth}
\resizebox{0.95\columnwidth}{!}{
\begin{tikzpicture}[thick, font=\scriptsize]
% Left column - inputs
% Input nodes

\draw[decorate,decoration={brace,amplitude=10pt,mirror}]
        (-1.6,1.8) -- (-1.6,-1.0) node [midway,xshift=-1cm, align=center] {SAA \\under \\ Scenario $i$};

\node (data_transit) at (0,0) {%
  \begin{tabular}{c}
    \faUsers \\[2pt]
    6,659 transit trips \\
    in Dataset \textsc{L-Arc-8}
  \end{tabular}
};

\node (data_drive) at (0, 1.4) {%
  \begin{tabular}{c}
    \faUsers \\[2pt]
    31,520 driving trips \\
    in Dataset \textsc{L-Arc-8}
  \end{tabular}
};

\node at (1.6, 1.4) {$+$};

% Apply C_core
\node (adopting_trips) at (3, 1.4) {%
  \begin{tabular}{c}
    \textcolor{blue}{\faUsers} \\[2pt] \textcolor{blue}{Transit trips} \\
    \textcolor{blue}{with other choices}
  \end{tabular}
};
\node (core_trips) at (3, -0.8) {%
  \begin{tabular}{c}
    \textcolor{blue}{\faUsers} \\[2pt] \textcolor{blue}{Core trips $\mathcal{T}_{\text{core}}^i$}
  \end{tabular}
};

% Path sets
% Path sets
\node (latent_paths) at (7.2, 1.4) {%
  \begin{tabular}{c}
    \textcolor{blue}{\faWalking\,\faBus\,\faWalking} \\[2pt]
    \textcolor{blue}{Paths of $\mathcal{T}_{\text{latent}}^i$} \\
    \textcolor{blue}{\&} \\  \textcolor{blue}{Adoptions $d_{r,p}^i$}
  \end{tabular}
};
\node (core_paths) at (7.2, -0.8) {%
  \begin{tabular}{c}
     \textcolor{blue}{\faWalking\,\faBus \faBus\,\faTrain} \\[2pt] \textcolor{blue}{Paths of $\mathcal{T}_{\text{core}}^i$}
  \end{tabular}
};

% Single arrow that splits
\draw[->] (data_transit) -| node[above, pos=0.25] {\scriptsize Apply $\mathcal{C}_{\mathrm{core}}$} (core_trips);
\draw[->] (data_transit) -| (adopting_trips);

\node[draw=blue, dashed, thick, rounded corners, fit=(adopting_trips)(data_drive), inner sep=0.15cm, label={[anchor=south, text=blue, font=\scriptsize]north: \faUsers\ Latent Trips $\mathcal{T}_{\text{latent}}^i$}] (latent_box) {};

% Arrows to paths
\draw[->] (latent_box) -- node[above, align=center, pos=0.6] {\scriptsize Collect\\ \scriptsize paths} node[below, align=left, pos=0.6] {\scriptsize Apply $\mathcal{C}_{\text{adopt}}$} (latent_paths);

\draw[->] (core_trips) -- node[above, align=center, pos=0.6] {\scriptsize Collect \\  \scriptsize paths} (core_paths);
\end{tikzpicture}
}
\end{subfigure}
%%%%%%%%%%%%
\vskip\baselineskip
\begin{subfigure}[b]{0.99\columnwidth}
\resizebox{0.9\columnwidth}{!}{
\begin{tikzpicture}[
  line width=1.2pt,
  font=\small,
  box/.style={rectangle, draw, rounded corners, minimum width=4cm, minimum height=1cm, align=center}
]

\node[box] (box2) at (0, - 1.2) {SAA with $I'$ Scenarios}; %{$I'$ (1000) SAA Instances};

\node[box] (box1) at (0,0) {SAA with $I$ Scenarios}; %{$I$ (50) SAA Instances};

\node (cp_model) at (5, 0) {
  \begin{tabular}{c}
    \faDesktop \\[2pt] CP Model
  \end{tabular}
};
\node (network) at (9, 0) {
  \begin{tabular}{c}
    \faProjectDiagram  \\[2pt] A Network Design
  \end{tabular}
};

\draw[->] (box1) -- (cp_model);
\draw[->] (cp_model) -- node[above] { solving} (network);
\draw[->] (network.south) |- node[above, pos=0.75] { evaluation with additional SAA scenarios} (box2.east);

\draw [decorate,decoration={brace,amplitude=10pt,mirror}]
        (-2.5,0.5) -- (-2.5,-1.5) node [midway,xshift=-0.8cm, align=center] {CP \\ Part};

\end{tikzpicture}
}
\end{subfigure}

\caption{A Flowchart showing the Atlanta case study within the 2LRC-TND framework. Blue indicates that these components change in each SAA scenario.}
\label{fig:flowchart}
\vspace{-5mm}
\end{figure}

%% file: sections/5_Results.tex
\section{Computational Results}
\vspace{-1.5mm}
This section begins by presenting the machine learning results from the two choice models. It then reports the optimization results of the 2LRC-TND framework applied to the aforementioned case study.
\vspace{-3.5mm}
\subsection{Results on Choice Modeling}
\vspace{-2mm}
\begin{table}[!t]
\centering
\caption{Choice modeling results averaged over 100 runs (100 different training-test split), with standard deviations in parentheses.}
\resizebox{0.95\columnwidth}{!}{
\begin{tabular}{l r r r  r r r}
\toprule
& \multicolumn{3}{c}{$\mathcal{C}_\mathrm{core}$ (on \textsc{Survey-Arc} Dataset)} & \multicolumn{3}{c}{$\mathcal{C}_{adopt}$ (on \textsc{L-Arc-9} Dataset)} \\
\cmidrule(lr){2-4} \cmidrule(lr){5-7}
\makecell{Models} & \makecell{F1-score} & \makecell{Accuracy} & \makecell{AUC-ROC}
                & \makecell{Weighted F1} & \makecell{Accuracy} & \makecell{AUC-ROC} \\
\cmidrule(lr){1-1} \cmidrule(lr){2-4} \cmidrule(lr){5-7}
L   & 0.844 (0.005) & 0.780 (0.006) & 0.805 (0.008)
    & 0.912 (0.003) & 0.924 (0.004) & 0.820 (0.015) \\
DNN & 0.866 (0.015) & 0.803 (0.016) & 0.799 (0.011)
    & 0.907 (0.022) & 0.897 (0.036) & 0.807 (0.029) \\
RF  & 0.885 (0.003) & 0.821 (0.005) & 0.819 (0.008)
    & 0.923 (0.003) & 0.933 (0.003) & 0.855 (0.015) \\
\bottomrule
\end{tabular}
}
\label{tab:ml_results}
%\vspace{-1mm}
\end{table}

Table~\ref{tab:ml_results} summarizes the performance of choice models on classification tasks. While the logit model serves as a baseline due to its simplicity, it still produces reasonably consistent results. Tree-based models, such as RF, outperform the other two models. In contrast, DNNs demonstrate comparatively weaker performance across both tasks. This under-performance is likely attributable to the tabular nature of the data, a context in which DNNs usually struggle. For RF's hyper-parameters, the best configuration for $\mathcal{C}_\mathrm{core}$ is 2000 estimators with a maximum depth of 25. For $\mathcal{C}_{adopt}$, RF uses 1000 estimators, a maximum depth of 10, and a class weighting ratio of 1:5 between class 0 and class 1 during training.

Overall, the results are consistent and exhibit low standard deviation across 100 runs. For $\mathcal{C}_{adopt}$, the weighted F1-score is employed to address the highly imbalanced nature of the dataset. Given this imbalance, models achieving a AUC-ROC score above 0.8 can be considered strong. Despite the classification performance, the models also yield valuable insights that usually surpass the capabilities of basic rule-based methods, and they are particularly useful when integrated into larger optimization frameworks. These results further emphasize the need for more comprehensive and balanced data collection to better capture the complexities inherent in both tasks.
\vspace{-3.5mm}
\subsection{Results on 2LRC-TND}
\begin{figure}[t!]
    \centering
    \begin{subfigure}[b]{0.23\columnwidth}
        \centering
        \includegraphics[width=\textwidth]{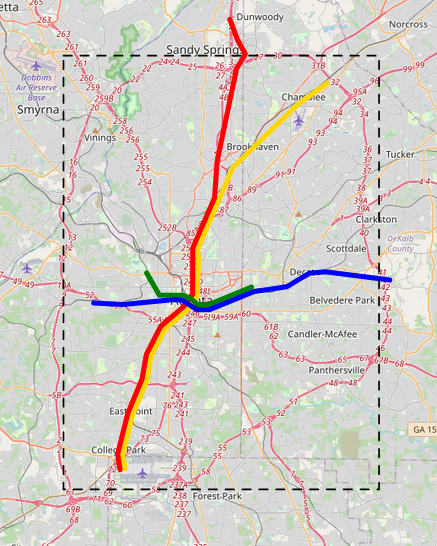}
        \caption{Rail System \& \\  Rectangular Area}
        \label{subfig:rail_and_rectangle}
    \end{subfigure}
    \hfill
    \begin{subfigure}[b]{0.23\columnwidth}
        \centering
        \includegraphics[width=\textwidth]{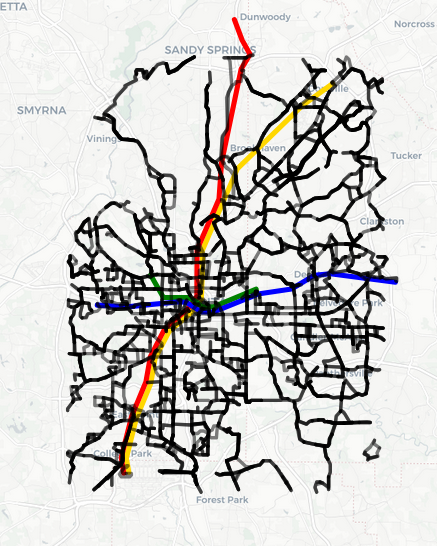}
        \caption{FD \\ ($B$ = 30k)}
        \label{fig:design_0}
    \end{subfigure}
    \hfill
    \begin{subfigure}[b]{0.23\columnwidth}
        \centering
        \includegraphics[width=\textwidth]{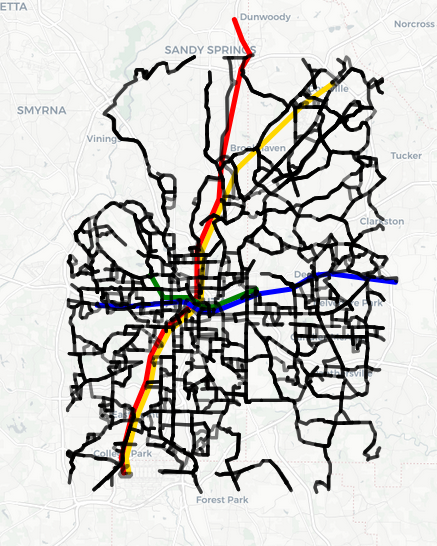}
        \caption{2LNC-TND \\ ($B$ = 30k)}
        \label{fig:design_1}
    \end{subfigure}
    \hfill
    \begin{subfigure}[b]{0.23\columnwidth}
        \centering
        \includegraphics[width=\textwidth]{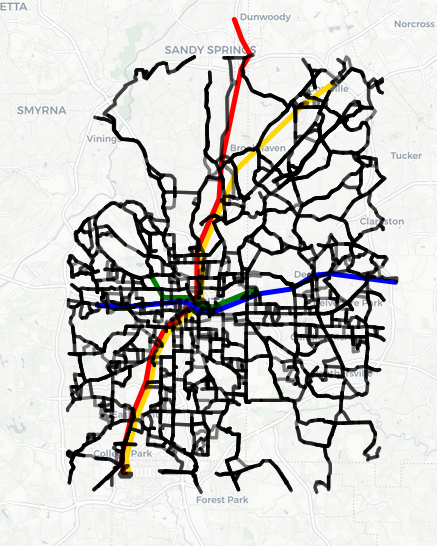}
        \caption{2LNC-TND \\ ($B$ = 35k)}
        \label{fig:design_2}
    \end{subfigure}
    \caption{Selected Maps Related to the Case Study. (Designs are Hypothetical)}
    \label{fig:four_figures}
\vspace{-3mm}
\end{figure}

\vspace{-2mm}
Table~\ref{tab:result} summarizes the computational results on optimization. Random Forest is employed as $\mathcal{C}_\mathrm{core}$ and $\mathcal{C}_{adopt}$ in 2LRC-TND because its simplicity for implementation (compared to DNNs) and superior performance in prediction results demonstrated in prior experiments. Selected visualizations (due to page limit) of  designs are shown in Figures~\ref{fig:four_figures}. Examining the network designs, it appears that all runs open a similar number of arcs within the same budget. This outcome occurs because the model tries to utilize the entire available budget. The resulting networks largely preserve existing MARTA corridors while reallocating capacity toward underserved areas with high latent demand.

\begin{table*}[!t]
    \centering
    \caption{2LRC-TND results. For evaluations, the reported metrics are averaged across senarios: 50 ($I$) of them are used in solving and 1,000 ($I'$) only for additional evaluation. For the Adopted\% metric, the denominator is $|\mathcal{T}_{latent}|$.}
    \resizebox{0.99\textwidth}{!}{
    \begin{tabular}{l l r r   r r r r r r  r }
    \toprule
    %%%%%%%%%%%%%%%%%%%%%% %%%%%%%%%%%%%%%%%%%%%% %%%%%%%%%%%%%%%%%%%%%%
    & & & & \multicolumn{6}{c}{\textbf{Averaged Results over 50 ($I$) SAA Senarios}} & \multicolumn{1}{c}{\textbf{Eval. with $I'$}} \\
    \cmidrule(lr){5-10} \cmidrule(lr){11-11}
    %%%%%%%%%%%%%%%%%%%%%% %%%%%%%%%%%%%%%%%%%%%% %%%%%%%%%%%%%%%%%%%%%%
      \makecell[l]{\textbf{Budget} \\ (\$)}
    & \makecell[l]{\textbf{Run}}
    & \makecell{\textbf{Run} \\ \textbf{Time} \\ \textbf{(hours)}}
    & \makecell[r]{\textbf{\#} \textbf{Opened} \\ \textbf{Bus} \\ \textbf{Arcs}}
    & \makecell{\textbf{\#} \\ \textbf{Core} \\ $|\mathcal{T}_{core}|$}
    & \makecell{\textbf{\#} \\ \textbf{Latent} \\ $|\mathcal{T}_{latent}|$}
    & \makecell{\textbf{\# (\%)} \\ \textbf{Adopted}}
    & \makecell{\textbf{Coverage}}
    & \makecell[r]{\textbf{Coverage} \\ \textbf{Bound}}
    & \makecell[r]{\textbf{ Travel} \\ \textbf{Time} \\  \textbf{(min)}}
    & \makecell{$\mathbf{Eval(\mathbf{z})}$ \\ \textbf{(\# Violated)}} \\
    \midrule
    % %%%%%%%%%%%%%%%%%%%%%% %%%%%%%%%%%%%%%%%%%%%% %%%%%%%%%%%%%%%%%%%%%% %%%%%%%%%%%%%
    \multirow{4}{*}{30K}
                         & FD & 0.04 & 2824 & - & - & - & 13086 & - & 57.13 & -  \\
                         & Naive RB & 2.23 & 2844 & 2139 & 36040 &  5676 (15.75\%) &  7814 &  8819 & 56.80 & 7809.15 \phantom{0}(0) \\
                         & Deterministic & 0.03 &  2701 & 4776& 33403 & 966 \phantom{0}(2.89\%) & 5742 & 5742 & 53.15 & 13338.58 (15) \\
                         & 2LRC-TND & 12.00 & 2871 & 4259 & 33920 & 11132 (32.82\%) & 15391 & 16655 & 56.44 & 15390.11 \phantom{0}(0) \\
    \midrule
    %%%%%%%%%%%%%%%%%%%%%% %%%%%%%%%%%%%%%%%%%%%% %%%%%%%%%%%%%%%%%%%%%%
    \multirow{4}{*}{35K}
                        & FD & 0.05 & 3148 & - & - & - & 13718 & - & 57.10 & -  \\
                        & Naive RB &  1.98 & 3319 & 2139 & 36040 &  6182 (17.15\%) &  8321 &  8819 & 57.03 & 8313.00  \phantom{0}(0) \\
                        & Deterministic & 0.03 &  2838 & 4776 & 33403 & 966 \phantom{0}(2.89\%) & 5742 & 5742 & 53.12 & 13474.61 (15) \\
                        & 2LRC-TND &  3.88 & 3313 & 4259 & 33920 & 11745 (34.63\%) & 16005 & 16655 & 56.56 & 15999.53 \phantom{0}(0) \\
    \bottomrule
    \end{tabular}
    }
    \label{tab:result}
\end{table*}

The first benchmark, FD, assumes no demand uncertainties and solves the CP model using only the existing 6,659 transit trips. However, after the network is designed, $\mathcal{C}_{\mathrm{core}}$ and $\mathcal{C}_{\mathrm{adopt}}$ (both using the random forest models in 2LRC-TND) are applied to evaluate the actual number of users who would adopt the service, with results reported in the ``coverage'' column in the table. Without considering demand uncertainties (run FD), the network design will still attract riders to join the system, and some riders among the 6,659 will reject it. The new design by FD can still attract some riders because the objective does not aim at budget savings, resulting in a quite extensive network. In contrast, 2LRC-TND accounts for demand uncertainties and achieves a much higher coverage by approximately 20\%. These results demonstrate that through the redesign of 2LRC-TND, the agency can attract additional riders by offering services to those who previously lacked transit access. The column coverage bound indicates the maximum ridership that an agency can cover (assuming no budget limit). This value is determined prior to solving the CP model, during pre-processing.

The second benchmark, Naive Rule-based (RB), uses nested if-else conditions based on travel time and origin-destination locations to generate probabilities, serving the same purpose as $\mathcal{C}_{\mathrm{core}}$ and $\mathcal{C}_{\mathrm{adopt}}$ in 2LRC-TND but without machine learning. The first observation is, using the rule-based models, the CP can be solved optimally within a short time, whereas 2LRC-TND ($B$ = 30K) only reports the best feasible solution after the run time limit of 12 hours is reached. This is likely due to the machine learning model predicting a greater number of core riders, which requires the CP model to provide a path for each core rider, thereby increasing computational difficulties. Secondly, although the core riders differ across runs, all experiments demonstrate that 2LRC-TND achieves a notable number of adopted riders. Specifically, RF yields a higher adoption rate by learning directly from data, while Naive RB depends on manually defined rules that lack consideration of geographical and demographic information.

To demonstrate the importance of stochastic modeling, a third benchmark employing deterministic choice models is evaluated against 2LRC-TND. This configuration utilizes the same machine learning models as 2LRC-TND; however, they generate deterministic binary decisions using a typical fixed threshold of 0.5, thus eliminating the need for SAA scenarios in $I$. In this case study, since the predicted probability of adopting a path for latent trips is typically below 0.5 in most cases, this deterministic approach underestimates potential ridership adoption. As shown in Table~\ref{tab:result}, the deterministic approach achieves a significantly weaker adoption rate across both budget levels.

For core riders and adopted riders using the final network design, Table~\ref{tab:result} also reports their travel times. It should be noted again that the model itself does not assign specific paths to riders. It ensures that core riders have at least one feasible path in $P_{r}$ and also attempts to increase the number of feasible paths for latent riders. Once the network design is established, the fastest path can be selected for riders, since this approach is intuitive.

Lastly, the designs are evaluated using 1000 ($I'$) additional SAA scenarios. Results show that all core riders are served under the respective designs with no violation in most of the cases, which is likely because each $\mathcal{P}_r$ contains four paths and the fixed rail arcs are effectively utilized by core riders. However, Deterministic case exhibits 15 constraint violations when evaluated on these scenarios indicating the need for the stochastic decision-making models. Additionally, the $\mathbf{Eval(\mathbf{z})}$ results closely match the coverage values from the evaluation with $I$ for the 2LRC-TND approach, highlighting that the results are highly stable after solving from only 50 SAA scenarios.

%% file: sections/6_Conclusion.tex
\vspace{-3.5mm}
\section{Conclusion}
\vspace{-3.5mm}
This paper introduces the 2LRC-TND framework to address the limitations of the traditional transit network design approaches that assume fixed demand. Leveraging CSO, the framework integrates machine learning-based choice models and constraint programming-based optimization models to produce transit network designs. A large-scale case study over 38,000 trips from Atlanta demonstrates that 2LRC-TND can improve ridership within reasonable budget limits, highlighting its practicality and potential social impact.
Several promising directions for future research include design of more targeted surveys for enhancing data collection and extension of 2LRC-TND to emerging transportation systems.

%% file: reference.bib
@article{kleywegt2002sample,
  title={The sample average approximation method for stochastic discrete optimization},
  author={Kleywegt, Anton J and Shapiro, Alexander and Homem-de-Mello, Tito},
  journal={SIAM Journal on Optimization},
  volume={12},
  number={2},
  pages={479--502},
  year={2002},
  publisher={SIAM}
}

@article{guan2024path,
  title={Path-based formulations for the design of on-demand multimodal transit systems with adoption awareness},
  author={Guan, Hongzhao and Basciftci, Beste and Van Hentenryck, Pascal},
  journal={INFORMS Journal on Computing},
  year={2024},
  volume={36},
  number={6},
  pages={1359-1756},
  publisher={Informs}
}

@InProceedings{
    Basciftci2020,
    author="Basciftci, Beste
    and Van Hentenryck, Pascal",
    editor="Hebrard, Emmanuel
    and Musliu, Nysret",
    title="Bilevel Optimization for On-Demand Multimodal Transit Systems",
    booktitle="Integration of Constraint Programming, Artificial Intelligence, and Operations Research",
    year="2020",
    pages="52--68",
    publisher="Springer International Publishing",
}

@article{sadana2025survey,
  title={A survey of contextual optimization methods for decision-making under uncertainty},
  author={Sadana, Utsav and Chenreddy, Abhilash and Delage, Erick and Forel, Alexandre and Frejinger, Emma and Vidal, Thibaut},
  journal={European Journal of Operational Research},
  volume={320},
  number={2},
  pages={271--289},
  year={2025},
  publisher={Elsevier}
}

@inproceedings{guan2025contextual,
  title={Contextual stochastic optimization for school desegregation policymaking},
  author={Guan, Hongzhao and Gillani, Nabeel and Simko, Tyler and Mangat, Jasmine and Van Hentenryck, Pascal},
  booktitle={Proceedings of the AAAI Conference on Artificial Intelligence},
  volume={39(27)},
  pages={28024--28032},
  year={2025}
}

@article{ye2024contextual,
  title={Contextual Stochastic Optimization for Omnichannel Multi-Courier Order Fulfillment Under Delivery Time Uncertainty},
  author={Ye, Tinghan and Cheng, Sikai and Hijazi, Amira and Van Hentenryck, Pascal},
  journal={Manufacturing \& Service Operations Management},
  year={2025}
}

@article{bertsimas2020predictive,
  title={From predictive to prescriptive analytics},
  author={Bertsimas, Dimitris and Kallus, Nathan},
  journal={Management Science},
  volume={66},
  number={3},
  pages={1025--1044},
  year={2020},
  publisher={INFORMS}
}

@article{basciftci2023capturing,
  title={Capturing travel mode adoption in designing on-demand multimodal transit systems},
  author={Basciftci, Beste and Van Hentenryck, Pascal},
  journal={Transportation Science},
  volume={57},
  number={2},
  pages={351--375},
  year={2023},
  publisher={INFORMS}
}

@misc{arc2020survey,
  author        = {{Atlanta Regional Commision}},
  title         = {Regional On-Board Transit Survey 2019 Final Report},
  year          = {2020},
  howpublished  = {\url{https://cdn.atlantaregional.org/wp-content/uploads/final-report-arc-2019-regional-transit-on-board-survey.pdf}},
  note          = {Accessed: 2025-06-15}
}

@article{auad2021resiliency,
  title={Resiliency of on-demand multimodal transit systems during a pandemic},
  author={Auad, Ramon and Dalmeijer, Kevin and Riley, Connor and Santanam, Tejas and Trasatti, Anthony and Van Hentenryck, Pascal and Zhang, Hanyu},
  journal={Transportation Research Part C: Emerging Technologies},
  volume={133},
  pages={103418},
  year={2021},
  publisher={Elsevier}
}

@article{farahani2013review,
title = {A review of urban transportation network design problems},
journal = {European Journal of Operational Research},
volume = {229},
number = {2},
pages = {281-302},
year = {2013},
author = {Reza Zanjirani Farahani and Elnaz Miandoabchi and W.Y. Szeto and Hannaneh Rashidi},
}

@article{borndorfer2007column,
  title={A column-generation approach to line planning in public transport},
  author={Bornd{\"o}rfer, Ralf and Gr{\"o}tschel, Martin and Pfetsch, Marc E},
  journal={Transportation Science},
  volume={41},
  number={1},
  pages={123--132},
  year={2007},
  publisher={INFORMS}
}

@article{schobel2012line,
  title={Line planning in public transportation: models and methods},
  author={Sch{\"o}bel, Anita},
  journal={OR Spectrum},
  volume={34},
  number={3},
  pages={491--510},
  year={2012},
  publisher={Springer}
}

@article{maheo2019benders,
  title={Benders decomposition for the design of a hub and shuttle public transit system},
  author={Maheo, Arthur and Kilby, Philip and Van Hentenryck, Pascal},
  journal={Transportation Science},
  volume={53},
  number={1},
  pages={77--88},
  year={2019},
  publisher={INFORMS}
}

@article{bertsimas2021data,
  title={Data-driven transit network design at scale},
  author={Bertsimas, Dimitris and Ng, Yee Sian and Yan, Julia},
  journal={Operations Research},
  volume={69},
  number={4},
  pages={1118--1133},
  year={2021},
  publisher={INFORMS}
}

@inproceedings{klier2008line,
  title={Line optimization in public transport systems},
  author={Klier, Michael J and Haase, Knut},
  booktitle={Operations Research Proceedings 2007: Selected Papers of the Annual International Conference of the German Operations Research Society (GOR) Saarbr{\"u}cken, September 5--7, 2007},
  pages={473--478},
  year={2008},
  organization={Springer}
}

@article{klier2015urban,
  title={Urban public transit network optimization with flexible demand},
  author={Klier, Michael J and Haase, Knut},
  journal={Or Spectrum},
  volume={37},
  pages={195--215},
  year={2015},
  publisher={Springer}
}

@article{canca2016general,
  title={A general rapid network design, line planning and fleet investment integrated model},
  author={Canca, David and De-Los-Santos, Alicia and Laporte, Gilbert and Mesa, Juan A},
  journal={Annals of Operations Research},
  volume={246},
  number={1},
  pages={127--144},
  year={2016},
  publisher={Springer}
}

@article{cipriani2012transit,
  title={Transit network design: A procedure and an application to a large urban area},
  author={Cipriani, Ernesto and Gori, Stefano and Petrelli, Marco},
  journal={Transportation Research Part C: Emerging Technologies},
  volume={20},
  number={1},
  pages={3--14},
  year={2012},
  publisher={Elsevier}
}

@article{guihaire2008transit,
  title={Transit network design and scheduling: A global review},
  author={Guihaire, Val{\'e}rie and Hao, Jin-Kao},
  journal={Transportation Research Part A: Policy and Practice},
  volume={42},
  number={10},
  pages={1251--1273},
  year={2008},
  publisher={Elsevier}
}

@article{xie2003work,
  title={Work travel mode choice modeling with data mining: decision trees and neural networks},
  author={Xie, Chi and Lu, Jinyang and Parkany, Emily},
  journal={Transportation Research Record},
  volume={1854},
  number={1},
  pages={50--61},
  year={2003},
  publisher={SAGE Publications Sage CA: Los Angeles, CA}
}

@article{zhao2020prediction,
  title={Prediction and behavioral analysis of travel mode choice: A comparison of machine learning and logit models},
  author={Zhao, Xilei and Yan, Xiang and Yu, Alan and Van Hentenryck, Pascal},
  journal={Travel behaviour and society},
  volume={20},
  pages={22--35},
  year={2020},
  publisher={Elsevier}
}

@article{lee2018comparison,
  title={Comparison of four types of artificial neural network and a multinomial logit model for travel mode choice modeling},
  author={Lee, Dongwoo and Derrible, Sybil and Pereira, Francisco Camara},
  journal={Transportation Research Record},
  volume={2672},
  number={49},
  pages={101--112},
  year={2018},
  publisher={SAGE Publications Sage CA: Los Angeles, CA}
}

@article{ma2020travel,
  title={Travel mode choice prediction using deep neural networks with entity embeddings},
  author={Ma, Yixuan and Zhang, Zhenji},
  journal={IEEE Access},
  volume={8},
  pages={64959--64970},
  year={2020},
  publisher={IEEE}
}

@article{bertsimas2016inventory,
  title={Inventory management in the era of big data},
  author={Bertsimas, Dimitris and Kallus, Nathan and Hussain, Amjad},
  journal={Production and Operations Management},
  volume={25},
  number={12},
  pages={2006--2009},
  year={2016},
  publisher={SAGE Publications Sage CA: Los Angeles, CA}
}

@article{wang2025data,
  title={Data-driven ordering policies for target oriented newsvendor with censored demand},
  author={Wang, Wanpeng and Deng, Shiming and Zhang, Yuying},
  journal={European Journal of Operational Research},
  volume={323},
  number={1},
  pages={86--96},
  year={2025},
  publisher={Elsevier}
}

@article{jiang2025optimising,
  title={Optimising hurricane shelter locations with smart predict-then-optimise framework},
  author={Jiang, Zhenlong and Ji, Ran},
  journal={International Journal of Production Research},
  volume={63},
  number={8},
  pages={2905--2925},
  year={2025},
  publisher={Taylor \& Francis}
}

@article{tang2025enhanced,
  title={Enhanced route planning with calibrated uncertainty set},
  author={Tang, Lingxuan and Luo, Rui and Zhou, Zhixin and Colombo, Nicolo},
  journal={Machine Learning},
  volume={114},
  number={5},
  pages={1--16},
  year={2025},
  publisher={Springer}
}

@Misc{arc2025activity,
  author        = {{Atlanta Regional Commision}},
  title        = {{Activity Based Model}},
  year         = {2025},
  howpublished  = {\url{https://abmfiles.atlantaregional.com/}},
  note          = {Accessed: 2025-07-25}
}

@Article{guan2022heuristic,
  author  = {Guan, Hongzhao and Basciftci, Beste and Van Hentenryck, Pascal},
  journal = {Transportation Science},
  title   = {{Bilevel Optimization and Heuristic Algorithms for Integrating Latent Demand into the Design of Large-Scale Transit Systems}},
  year    = {2026},
}
